\documentclass{amia}

\usepackage{svg, todonotes, boldline, bbm, makecell, multirow, wrapfig, adjustbox}
\usepackage{graphicx, boldline, makecell}
\graphicspath{{./figures/}}
\definecolor{green}{RGB}{24, 135, 50}
\definecolor{darkred}{RGB}{125, 14, 6}
\definecolor{blue}{RGB}{19, 19, 237}
\definecolor{darkblue}{RGB}{8, 6, 105}
\definecolor{grey}{RGB}{91, 91, 94}

\usepackage[labelfont=bf]{caption}
\usepackage[superscript,nomove]{cite}
\usepackage{color}
\usepackage{todonotes}
\usepackage{boldline, makecell}
\usepackage{amsmath, amssymb, url, bbm}
\usepackage[T1]{fontenc}

\usepackage[symbol]{footmisc}

\begin{document}

\title{Weakly Supervised Classification of Vital Sign Alerts as Real or Artifact}

\author{Arnab Dey$^{1,2}$, Mononito Goswami, B.Tech.$^{1}$, Joo Heung Yoon, MD$^{3}$, Gilles Clermont, MD$^{3}$, Michael Pinsky, MD$^{3}$, Marilyn Hravnak, PhD, RN$^{3}$, Artur Dubrawski, PhD$^{1}$}

\institutes{
  \textsuperscript{\rm 1} Auton Lab, School of Computer Science, Carnegie Mellon University, Pittsburgh, PA; \textsuperscript{\rm 2}~Wallace H. Coulter Department of Biomedical Engineering; Georgia Institute of Technology, Atlanta, GA; \textsuperscript{\rm 3}~School of Medicine, University of Pittsburgh, Pittsburgh, PA \\
}
\maketitle

\noindent{\bf Abstract}

\textit{A significant proportion of clinical physiologic monitoring alarms are false. This often leads to alarm fatigue in clinical personnel, inevitably compromising patient safety. To combat this issue, researchers have attempted to build Machine Learning (ML) models capable of accurately adjudicating Vital Sign (VS) alerts raised at the bedside of hemodynamically monitored patients as real or artifact. Previous studies have utilized supervised ML techniques that require substantial amounts of hand-labeled data. However, manually harvesting such data can be costly, time-consuming, and mundane, and is a key factor limiting the widespread adoption of ML in healthcare (HC). Instead, we explore the use of multiple, individually imperfect heuristics to automatically assign probabilistic labels to unlabeled training data using weak supervision. Our weakly supervised models perform competitively with traditional supervised techniques and require less involvement from domain experts, demonstrating their use as efficient and practical alternatives to supervised learning in HC applications of ML.}

\section*{Introduction}
\label{sec:intro}


Intensive care patients who are at risk of cardiorespiratory instability (CRI) undergo continuous monitoring of vital sign (VS) parameters such as electrocardiography, plethysmography, pulse oximetry, and impedance pneumography. Recent advances in commercial bedside monitoring devices have made the sustained tracking of the physical state and health of a connected patient a real possibility. Without these devices, it would be practically impossible for medical practitioners to continually and attentively observe fast-evolving and heterogeneous VS parameters. However, even modern commercial devices have surprisingly inadequate support for identifying abnormal physiological variables in the form of simple exceedances of pre-determined normality thresholds \cite{otero2009addressing}. Moreover, it is not uncommon for patients to have atypical VS parameters due to occasional movement, electrical interference, or loose sensors \cite{chen}. 

\begin{wrapfigure}[16]{r}{0.5\textwidth}
    \centering
    \adjustbox{trim=0cm 0.8cm 0cm 0.5cm}{%
    \includegraphics[width=0.5\textwidth]{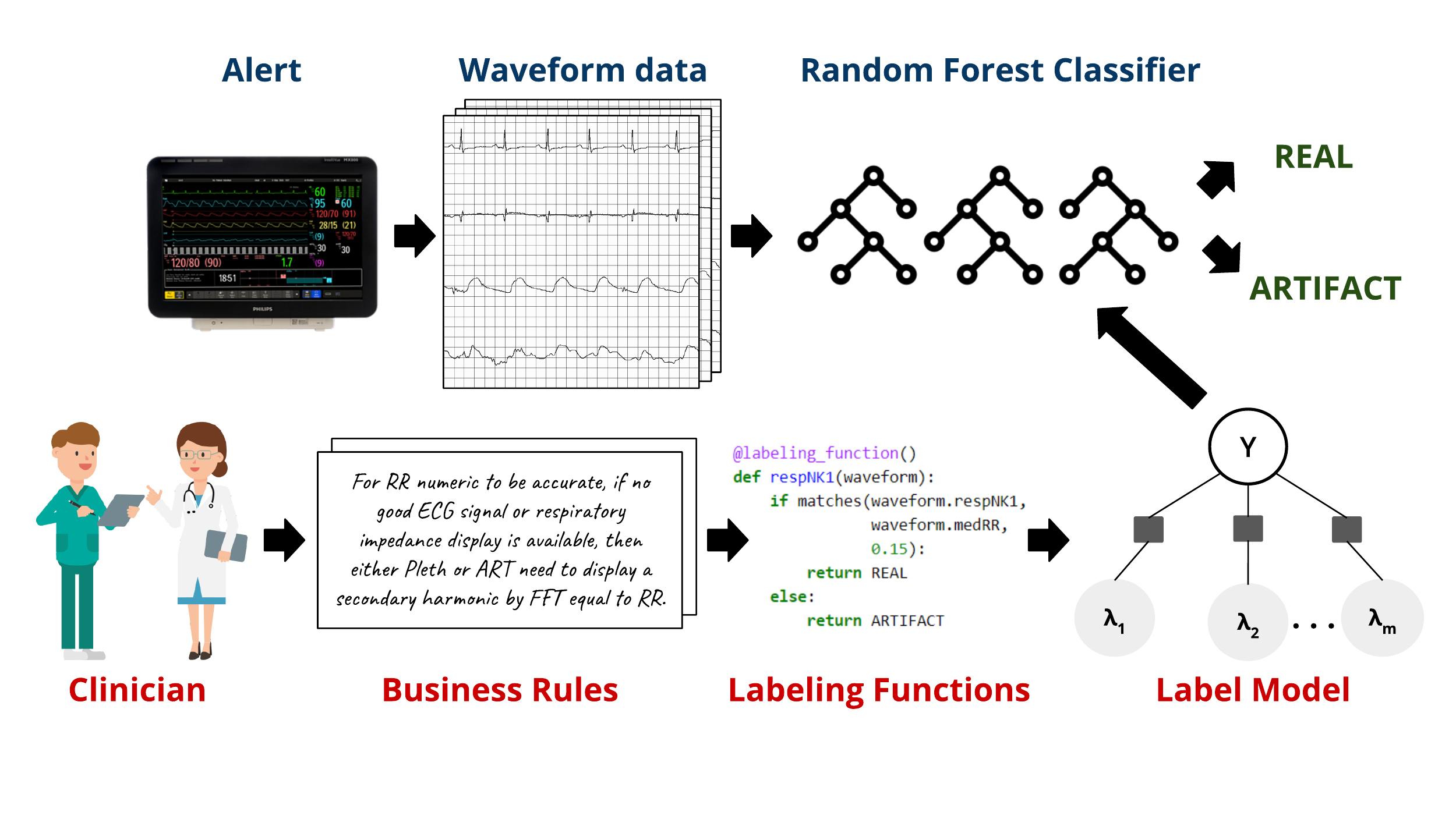}
    }
    \caption{Weak supervision pipeline for the binary classification of vital sign alerts. Heuristics given by domain experts are encoded into labeling functions whose votes are fed into a generative label model. This model then outputs probabilistic labels that are used for training a downstream real vs.\ artifact alert Random Forest classifier}
    \label{fig:WSPipeline}
\end{wrapfigure}

Indeed, numerous studies have shown a large percentage of these VS alerts to be false, or more formally, artifact \cite{sendelbach} - either of mechanical, electrical, or physiological nature \cite{chen, takla}. Additionally, medical practitioners may be exposed to up to 1,000 alarms per Intensive Care Unit (ICU) shift~\cite{ruskin}. The sheer amount of alarms in tandem with the high rate of artifacts can quickly lead to alarm desensitization and burnout in healthcare professionals. Multiple studies have concluded that the resulting alarm fatigue can have severe negative consequences for patient safety, with several incidents resulting in preventable harm or even death of a subject~\cite{sendelbach,ruskin,hravnakAlarm}. Furthermore, studies have shown that medical practitioners unintentionally respond to noisy work environments created by the loud and frequent blaring of artifactual clinical alarms, by becoming less engaging and empathetic towards patients~\cite{lewandowska2020impact, wung2018sensory, vitoux2018perceptions}. 

In addition to the added stress placed on medical practitioners, frequent alerts can also lead to increased physiological stress in the patient, metabolic impairment, sleep disturbance and even death \cite{hravnakAlarm}. Moreover, these frequent artifacts preclude the continuous monitoring of VS in postoperative ward patients, leading to the early warning signs of impending cardiorespiratory arrest to go unnoticed and untreated until it is too late \cite{khanna2019postoperative}. Not counting these many lives lost, the U.S.\ Food and Drug Administration (FDA) has reported over $500$ alarm-fatigue related patient deaths in the short span of five years~\cite{ruskin}.

Previous attempts to combat alarm fatigue have relied on advancements in adaptive filtering or explored the use of various Machine Learning (ML) paradigms, particularly supervised and active learning \cite{chen,hravnakArchive}. However, these methodologies require varying quantities of manually and pointilistically labelled data. Labeling alerts as real or artifact is not only time-intensive, but also a laborious, expensive, and mundane task that pulls experienced clinicians away from their patients. Furthermore, traditional ML paradigms do not easily adapt to evolving clinical expertise and changing problem definitions due to their reliance on pointillisticaly annotated data which must be re-labeled to accommodate each such problem redefinition. For example, sepsis is one of the most sought-after clinical conditions to predict. However, with the constantly evolving definition of sepsis, the labeling process is frequently affected, causing many annotations to become inconsistent with current guidelines \cite{giacobbe2021early}.

As an alternative, Weak Supervision (WS) involves harvesting general heuristics that clinicians would normally use to label the data by hand, and collectively using them to probabilistically reconstruct the labels for even vast amounts of unlabeled reference data. The hope is that downstream models trained with such automatically annotated data would perform as well as the models trained on data labeled in a point-by-point fashion, while greatly reducing the human effort and time needed to develop such models. This allows clinicians to focus where it matters, while enabling the development of accurate and efficient ML models, paving the way for materializing a significant social impact in healthcare. Recent work suggests that the proposed WS methodology can indeed accomplish such goals in some HC applications \cite{saab2020weak, goswami2021weak}. 
In this paper, we demonstrate the potential utility of WS to adjudicate bedside alerts as real vs.\ artifact using high-density waveform VS data collected in intensive care settings.

\section*{Related Work}
\paragraph{Alarm Fatigue}
Alarm fatigue caused by high rates of artifact VS alerts is a widely-studied problem and a variety of techniques have been adopted to combat it in previous research. Most approaches fall into two main categories: ($1$) artifact reduction, and ($2$) artifact detection. The former approach attempts to reduce the number of artifact alerts produced through internal improvements within the vital sign monitors and other biosignal-measuring devices. Advancements in adaptive filtering and other techniques to reduce artifacts in real time within the monitor itself have been developed \cite{sinhal, graybeal, he2017novel, paul2000transform, marque2005adaptive, lu2009removing}. But, due to the wide ranges of signal frequency and the diverse nature and causes of artifact alerts~\cite{lee2011automatic,couceiro2012detection}, the problem of alarm fatigue still persists \cite{chen}. This work aims to tackle alarm fatigue and the high rates of artifact alarms through the latter approach, which focuses on post-measurement artifact detection and alert adjudication. 

\paragraph{Clinical Settings}
A large body of research has been produced on post-measurement artifact detection in the past, but most approaches either look at ambulatory settings or are in the context of wearable devices and smartphones -- settings which are fundamentally different from the acute care clinical setting due to differences in physiological states of subjects, data quality, rates of motion and noise artifact, amount and type of available data, a priori likelihood of artifacts, and primary differences in the types of artifact that need to be detected~\cite{bashar2018developing, pollreisz2019detection, shimazaki2014cancellation}. 

\paragraph{Machine Learning Paradigms}
Prior research on artifact detection strictly in the clinical setting has been conducted, but most papers combat alarm fatigue through the use of traditional ML pipelines such as fully-supervised (FS), active and federated learning \cite{chen, hravnakArchive, caldas}. These efforts yielded great strides in VS alert classification capabilities, but require substantial amounts of expert-annotated reference data to train efficient and accurate classifiers. A distinct lack of analysis remains on classifiers trained in data- and label-scarce environments using models expressly suited for this application. 
To the best of our knowledge, our work is the first to apply weak supervision to the problem of VS alert classification.


\section*{Methodology}
\paragraph{Problem Formulation}
Broadly, given an alert, our goal is to classify it as a real or artifact. We specify each alert as a $4$-tuple $\mathcal{A}_i = (\texttt{pid}, \ \tau, \ t, \ d)$, where \texttt{pid} is a unique alert ID, $\tau \in \{\texttt{RR},\ \texttt{SpO}_2\}$ is the alert type, $t$ and $d$ are the starting time and duration of the $i^{th}$ alert. We assume that each alert is associated with an unobserved true class label $y \in \{0, 1\}$, where $0$ denotes real and $1$ denotes an artifact; and that for the duration of the alert, we have access to time series data $\mathcal{T}_i$ which includes both waveforms such as electrocardiogram (\texttt{ECG}) leads II and III and numerics, such as heart rate (\texttt{HR}), potentially sampled at different frequencies. We aim to use clinical intuition and expert knowledge encoded in several heuristics to obtain labels to train an downstream classification model $\mathcal{M}$. We define each heuristic, alternatively called a labeling function (LF), denoted by $\lambda: \mathcal{T} \times \mathcal{A} \rightarrow \{-1, 0, 1\}$ directly on timeseries data. A LF either abstains \{$-1$\} or votes for a particular class \{$0, 1$\} given an alert $\mathcal{A}$ and its associated waveform data $\mathcal{T}$. While we do not expect  any individual LF to have perfect accuracy or recall, we do expect them to have better than random accuracy whenever they do not abstain from voting. Starting with $n$ alerts $\mathcal{X} = \{(\mathcal{A}_i, \mathcal{T}_i)\}_{1 \dots n}$ and $m$ labeling functions $\Lambda = \{\lambda_i\}_{i = 1 \dots m}$, our goal is to learn a label model $\mathcal{L}$ which assigns a probabilistic label $\hat p(y \mid \Lambda), \ y \in \{0, 1\}$ to each alert in $X$. 

The label model learns from the overlaps, conflicts and (optionally) dependencies between the LFs using a factor graph as shown in Fig.~\ref{fig:WSPipeline}. In this work, we assume the LFs to be independent given the true class label. While this assumption may not always stand, most prior work \cite{goswami2021weak, saab2020weak, fries2019weakly} has shown that this simple label model may work well in practice. Let $\mathbb{Y} = \{y_i\}^n$ denote the vector of unobserved ground truth labels, and let $\Lambda_{ij}$ be the vote of the $j^{th}$ LF on the $i^{th}$ data point. We then define LF accuracy and propensity as $\phi^{Acc}(\Lambda_{ij},y_i) \triangleq \mathbbm{1}\{\Lambda_{ij}=y_i\}$, and $\phi^{Lab}(\Lambda_{ij},y_i) \triangleq \mathbbm{1}\{\Lambda_{ij}\neq 0\}$, respectively. Following Ratner et al.\cite{ratner}, we define the model of the joint distribution of $\Lambda$ and $Y$ as:
\begin{align*}\label{eq:condind}
    p_{\theta}(\Lambda,Y) &= \frac{1}{Z_\theta}\exp ( \sum_{j=1}^m\sum_{i=1}^n ( \theta_j \phi^{Acc}(\Lambda_{ij},y_i) + \theta_{j+m} \phi^{Lab}(\Lambda_{ij},y_i)))
\end{align*}
where $Z_\theta$ is a normalizing constant and $\theta$ are the canonical parameters for the LF accuracy and propensity. We use Snorkel~\cite{ratner2017snorkel} to learn $\theta$ by minimizing the negative log marginal likelihood given the observed $\Lambda$. Finally, given a set of training alerts $\{x_1, \dots, x_n\}, \ x_i \in \mathcal{X}$ we want to train an end model classifier $\mathcal{M}: \mathcal{X} \rightarrow \mathcal{Y}$ such that $\mathcal{M}(x) = y$. 

\paragraph{Vital Sign Data}
In this work we use a large single-center database comprising of vital sign data of patients admitted to critical care units of a large tertiary care research and teaching hospital. The data was curated and de-identified at the institution, whose Institutional Review Board deemed this research did not qualify as human subjects research. Cardiorespiratory vital sign alert data consisting of a variety of waveforms and numerics were collected with the Philips IntelliVue MX800 Monitor from a mix of ICU and Step Down Unit (SDU) patients. The data comprised of approximately $367,464$ monitoring hours with around $80$ hours of data from each patient. Numerics, including respiratory rate (\texttt{RR}), \texttt{HR}, oxygen saturation (\texttt{SpO$_2$}), and telemetric oxygen saturation (\texttt{SpO$_2$T}) were sampled at 1 Hz. Waveform data, including \texttt{ECG} lead II and lead III, plethysmographs (\texttt{pleth}), telemetric plethysmographs (\texttt{plethT}), arterial pressure waveforms (\texttt{ART}) derived from an indwelling arterial catheter, and respiratory waveforms (\texttt{resp}) from impedance pneumography, were all sampled at various frequencies. \texttt{ECG} lead II and lead III were sampled at both 250 Hz and 500 Hz. \texttt{Pleth}, \texttt{plethT}, and \texttt{ART} were all sampled at 125 Hz, and the \texttt{resp} waveform was sampled at 62.5 Hz. 

\paragraph{Vital Sign Alert Events}
We determined both RR and SpO$_2$/SpO$_2$T vital sign alerts by analyzing the \texttt{RR} numeric and \texttt{SpO$_2$}/\texttt{SpO$_2$T} numeric, respectively, on $4$ factors: ($1$) \textit{duration} - at least $5$ minutes of the respective numeric data was present, ($2$) \textit{persistence} - at least $70\%$ of the numeric values exceeded respective thresholds (\textless\space $10$ breaths per minute or \textgreater\space $29$ breaths per minute for \texttt{RR} and \textless\space 90\% for \texttt{SpO$_2$}/\texttt{SpO$_2$T}), ($3$) \textit{tolerance} of 5 minutes suggesting that consecutive alerts \textless\space 5 minutes apart were combined, and ($4$) density expectation of $65\%$ of numeric values present at a $1$ Hz sampling frequency. These factors ensured that the VS alerts we analyzed contained continuous spaced anomalies with minimal interruption and were sufficiently long to have clinical relevance. Inspired by prior work by Chen et al.\cite{chen} and Hravnak et al.\cite{hravnakArchive}, we only used the first $3$ minutes of each alert event for both RR and SpO$_2$/SpO$_2$T alert classification. Additionally, we broke each $3$ minute alert window into three $1$ minute windows, primarily as a way to artificially boost the sample size. The ground truth label for each of the alert windows was assumed to be the same as that of the parent event. In the rest of this paper, RR or SpO$_2$/SpO$_2$T alerts refer to these $1$ minute alert event windows. We analyzed $648$ RR alerts (\textit{$216$ events}), comprising of $477$ real alerts and $171$ artifacts, and $621$ SpO$_2$/SpO$_2$T alerts (\textit{$207$ events}), comprising of $432$ real alerts and $189$ artifacts. Of these $621$ SpO$_2$/SpO$_2$T alerts, $183$ were telemetric alerts ($87$ real, $96$ artifact), and $438$ are non-telemetric ($345$ real, $93$ artifact).

\paragraph{Expert Knowledge Informing Alert Classification}
Manually classifying artifact VS alerts is an arduous, repetitive, yet sufficiently objective process, largely governed by a set of guiding principles or ``business rules'' based on visual distinction and clinical intuition \cite{hravnakArchive}. In this work, we utilized business rules which were developed during an iterative, multi-professional process of manual review and annotation of a subset of VS alerts by a committee of expert clinicians with decades of emergency-care experience. This review was followed by group discussions that involved adjudication and recognition of visual commonalities which were later translated into natural language rules upon consensus.

Most business rules are based on the apparent disagreement between numerics recorded by the monitor, and corresponding numerics derived from recorded waveform data. For instance, most business rules to distinguish between real and artifact \texttt{RR} alerts are based on discrepancy of observed \texttt{RR} and the \texttt{RR} numeric derived from the \texttt{resp}, \texttt{pleth}, \texttt{plethT}, and \texttt{ART} waveforms. In this study, however, we were unable to derive \texttt{RR} from the \texttt{plethT} and \texttt{ART} waveforms after finding a large portion of the data for these waveforms to be missing or incomplete.
Similarly, SpO$_2$/SpO$_2$T alerts are more likely to be artifacts when the observed \texttt{HR} does not match \texttt{HR} derived from the \texttt{pleth}/\texttt{plethT} waveforms. Our label model leveraged the overlaps and conflicts between labeling functions built on different core methodologies to probabilistically label training data. Some business rules compared the \texttt{HR} derived from ECG lead III to that computed from the \texttt{pleth} waveform, and another examined whether patients are experiencing tachypnea (rapid breathing, with \texttt{RR} \textgreater\space 20) during an oxygen saturation alert. To improve reliability, some business rules also checked whether \texttt{resp} and \texttt{pleth} waveforms were too low or displayed a lack of pulsatility. 

\paragraph{From Expert Knowledge to Labeling Functions}
\begin{wrapfigure}[14]{r}{0.5\textwidth}
    \centering
    \adjustbox{trim=0cm 0.2cm 0cm 0.4cm}{%
    \includegraphics[width=0.5\textwidth]{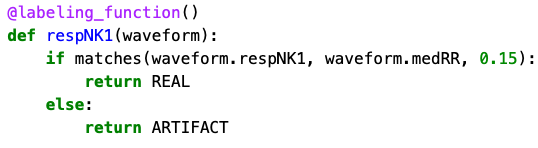}
    }
    \caption{This sample \texttt{RR} LF demonstrates the general design of these functions. Heuristics suggested by domain experts can be easily encoded as a set of simple conditional statements. In this specific case, when the value for the median \texttt{RR} derived from respiratory waveform data is within 15\% of the median \texttt{RR} numeric, the alert is labeled as real. Otherwise, it is labeled as artifact. 
    }
    \label{fig:exampleLF}
\end{wrapfigure}
Since most business rules relied on \texttt{RR} and \texttt{HR} numerics derived from the recorded waveforms, we developed multiple core methodologies with wide ranging accuracy, to compute these numerics.  For most business rules relying on derivations of \texttt{RR} and \texttt{HR}, it was important to be able to compute the primary/secondary harmonics and locate peaks in different waveforms. For instance, the \texttt{RR} closely corresponds to the median number of peaks and the primary harmonic of a clean \texttt{resp} waveform. We compute the former using a modified version of the Python \texttt{SciPy} package’s peak detection algorithm~\cite{virtanen} and extrema extraction algorithm proposed in Khodadad et al.\cite{khodadad2018optimized} as implemented in \texttt{Neurokit2}~\cite{makowski}. We computed the primary harmonic of the \texttt{resp} waveform by locating the highest peak of a periodogram modified by the Bohman windowing function. Prior to using the \texttt{resp} waveform, we processed it by linear detrending followed by a fifth order 2Hz low-pass IIR Butterworth filter~\cite{khodadad2018optimized}. 
To derive \texttt{RR} from the \texttt{pleth} and \texttt{ART} waveforms, we first processed them via a different, novel, multi-step methodology, which involved interpolating the tips of the peaks found using \texttt{SciPy}’s peak detection algorithm via spline interpolation. This was done to derive a new waveform designed to emulate the periodicity of the \texttt{resp} waveform, from which RR can be extracted via the same core methodologies.
For SpO$_2$/SpO$_2$T alerts, we derived the \texttt{HR} numeric from ECG lead II and III using the same core methodologies, after employing an ECG cleaning technique proposed in \texttt{Neurokit2}~\cite{makowski}.

Finally, we translated our business rules into labeling functions, building on the aforementioned core methodologies. As an example, Figure~\ref{fig:exampleLF} illustrates one such LF, comparing the observed median \texttt{RR} (\texttt{waveform.medRR}) with the \texttt{RR} derived from \texttt{resp} using the methods proposed in Khodadad et al.\cite{khodadad2018optimized} and implemented in \texttt{Neurokit2} (\texttt{waveform.respNK1}). We implemented a total of $8$ and $11$ noisy heuristics for the binary classification of RR alerts and SpO$_2$/SpO$_2$T alerts, respectively. 

\paragraph{From Labeling Functions to Alert Classifier}
We trained the label model $\mathcal{L}$ defined in the previous section using LFs for respective VS alerts, to obtain probabilistic labels for our training data. We used the label model implementation in Snorkel \cite{ratner2017snorkel} for the same. Samples not covered by any LF were filtered out, and the remaining probabilistic labels produced by the label model were translated into crisp binary training labels, which were then used to train a Random Forest (RF) model \cite{breiman2001random} to classify VS alerts as real versus artifact. RFs have been widely used in literature to learn complex decision boundaries for various classification problems \cite{goswami2020discriminating, goswami2020towards} and have also been shown to be effective for learning discriminative models of real versus artifact VS alert classification \cite{chen}. We trained RF models with $1000$ decision trees having a maximum depth of $5$ implemented using \texttt{scikit-learn} \cite{sklearn_api}.

\section*{Experimental setup}
\paragraph{Featurization}
In order to train the RF models, we utilized the features computed for use by our LFs such as the \texttt{RR} derived from a modified periodogram of the \texttt{resp} waveform (\texttt{respFFT}), wave amplitude of the \texttt{pleth} and \texttt{plethT} waveforms (\texttt{pulsatility} and \texttt{pulsatilityT}, respectively), etc., but we also extracted features from the raw waveforms and numerics themselves by computing a set of aggregate statistics (mean, standard deviation, kurtosis, skewness, median, 1st and 3rd quartile). For the RR alerts, we subsequently dropped the features calculated from the \texttt{ART}, \texttt{plethT}, and \texttt{ECG} lead III waveforms, and the \texttt{SpO$_2$T} numerics, due to more than $75\%$ of the alerts missing this data.  For SpO$_2$/SpO$_2$T alerts, we dropped features calculated from the \texttt{ART} waveform, for the same reason. Next, we replaced any missing values remaining in the data after incomplete features were removed with either a $0$ or $-1$ depending on the nominal ranges of the feature values. 


\paragraph{Baselines and Evaluation}
We compared our weakly supervised RF model (\texttt{Weak Sup.}) with its fully supervised counterpart trained using ground truth labels (\texttt{Fully Sup.}), probabilistic labels produced by the label model (\texttt{Prob. Labels}), and RF models trained using majority vote (\texttt{Majority Vote}) instead of the data programming label model. The majority vote model predicts what the majority of LFs voted for. All models were trained in a leave-one-patient-out (LOPO) cross-validation setting, where the models were trained on data from all but one patient, and tested on the held-out patient's data. This setting ensures that the models do not inadvertently fit to patient specific characteristics to prevent artificially inflating their performance. 

We compared all the models using a few different performance metrics including \texttt{accuracy} and \texttt{AUC}\footnote{The code for our experiments will be made publicly available at \url{www.github.com/--anonymous--} upon acceptance}. We also computed metrics of practical utility such as the false positive rate at $50\%$ true positive rate (\texttt{FPR 50\% TPR}), true positive rate at $1\%$ FPR (\texttt{TPR 1\% FPR}), etc. 
All models and LFs were implemented using Python programming language (version $3.8.1$), and experiments were carried out on a computing cluster with $64$ CPUs equipped with AMD Opteron $6380$ processors having a total of $252$ GB RAM. 

\paragraph{Additional Research Questions}
In addition to examining the efficacy of WS models for VS alert classification, we aimed to answer the following research questions. 

\noindent (1)~\textit{What patterns are our RF models learning?} Interpretability is important when ML models are deployed in clinical settings, especially when using complex models such RFs. We used \textit{Gini importance} (GI) \cite{breiman2001random} and \textit{permutation feature importance} (PFI) \cite{altmann2010permutation} to determine features which our weakly supervised model relied on the most while making label predictions (Figure ~\ref{fig:featureimp}). Since GI can be inflated for high-cardinality features, PFI was also analyzed to reliably understand feature importance, in line with prior work conducted in different settings \cite{goswami2020towards}. GI and PFI were evaluated by accessing the feature importance for a trained \texttt{scikit-learn} RF classifier, and using the \texttt{permutation\_feature\_importance} function in \texttt{scikit-learn}, respectively. 

\noindent (2)~\textit{How useful is the waveform data?} Since most of the previous work (e.g., Chen et al.\cite{chen} and Hravnak et al.\cite{hravnakArchive}) on VS alert classification did not utilize high-density VS waveform data, we were curious about the predictive utility of waveform data for classifying VS alerts. Consequently, we conducted ablation experiments by withholding waveform features while training and validating our weakly and fully supervised models using the same LOPO cross-validation procedure. However, we must note that the LFs informing the weakly supervised RF still had access to requisite waveform data, and therefore these experiments were not completely indicative of settings with a lack of waveform data.

\section*{Results}

\begin{table*}[!ht]
    \centering
    \resizebox{\textwidth}{!}{
    \begin{tabular}{c |c c c c c c|c c c c c c} \Xhline{1pt}
        & \multicolumn{6}{c |}{\textbf{Respiratory Rate Alerts}} & \multicolumn{6}{c}{\textbf{Oxygen Saturation Alerts}} \\
         & \texttt{Accuracy} & \texttt{AUC} & \texttt{FPR 50\% TPR} & \texttt{FNR 50\% TNR} & \texttt{TPR 1\% FPR} & \texttt{TNR 1\% FNR} & \texttt{Accuracy} & \texttt{AUC} & \texttt{FPR 50\% TPR} & \texttt{FNR 50\% TNR} & \texttt{TPR 1\% FPR} & \texttt{TNR 1\% FNR}\\ \hline
         \texttt{Weak Sup.} & \textbf{0.915} & 0.951 & 0.012 & 0.008 & 0.428 & \textbf{0.567} & 0.881 & 0.940 & \textbf{0.011} & 0.009 & 0.382 & \textbf{0.630}\\
         \texttt{Majority Vote} & 0.855 & \textbf{0.952} & \textbf{0} & 0.015 & 0.551 & 0.409 & 0.899 & 0.951 & \textbf{0.011} & \textbf{0.007} & \textbf{0.458} & \textbf{0.630}\\
         \texttt{Fully Sup.} & 0.886 & 0.898 & 0.07 & 0.023 & 0.038 & 0.304 & \textbf{0.903} & \textbf{0.964} & 0.016 & \textbf{0.007} & 0.345 & 0.582\\
         \texttt{Prob. Labels} & 0.894 & 0.936 & 0.006 &\textbf{ 0.002} & \textbf{0.577} & 0.550 & 0.844 & 0.902 & 0.016 & 0.037 & 0.151 & 0.143\\
         \texttt{WS w/o WF} & 0.887 & 0.918 & 0.035 & 0.010 & 0.031 & 0.474 & 0.709 & 0.754 & 0.111 & 0.197 & 0.012 & 0.032\\
         \texttt{Sup. w/o WF} & 0.838 & 0.871 & 0.053 & 0.019 & 0.004 & 0.146 & 0.730 & 0.825 & 0.037 & 0.106 & 0.002 & 0.243\\
         \texttt{Maj. w/o WF} & 0.792 & 0.899 & 0.07 & 0.019 & 0.080 & 0.199 & 0.702 & 0.779 & 0.058 & 0.167 & 0.025 & 0.069\\ \Xhline{1pt}
    \end{tabular}}
    \caption{We calculated various performance metrics of ML pipelines on the classification of RR \& SpO$_2$/SpO$_2$T alerts. Interestingly, we found the performance of the weakly supervised model to be comparable, and in some cases superior, to the fully supervised method for both alert types.}
    \label{tab:perfmetr2}
\end{table*}

\begin{wrapfigure}[17]{r}{0.5\textwidth}
\centering
    \resizebox{0.5\textwidth}{!}{
    \begin{tabular}{c c|c c|c c} \Xhline{1pt}
         & & \multicolumn{2}{c}{\textbf{GI}} & \multicolumn{2}{c}{\textbf{PFI}}\\ 
         & \texttt{Alert} & \texttt{WS} & \texttt{Fully Sup.} & \texttt{WS} & \texttt{Fully Sup.}\\ \hline
         \begin{tikzpicture} \draw[draw=white, fill=green, opacity=1.0] (0,0) -- (0,0.25) -- (0.25,0.25)-- (0.25,0) -- (0,0); \end{tikzpicture} & \multirow{5}{*}{\texttt{RR}} & std\_resp & q1\_rr & std\_rr & respHeight \\
        \begin{tikzpicture} \draw[draw=white, fill=green, opacity=0.8] (0,0) -- (0,0.25) -- (0.25,0.25)-- (0.25,0) -- (0,0); \end{tikzpicture} && respHeight &	respHeight 	&q1\_resp &	std\_resp\\
         \begin{tikzpicture} \draw[draw=white, fill=green, opacity=0.6] (0,0) -- (0,0.25) -- (0.25,0.25)-- (0.25,0) -- (0,0); \end{tikzpicture} && mean\_rr &	mean\_rr &	respFFT &	std\_rr\\
         \begin{tikzpicture} \draw[draw=white, fill=green, opacity=0.6] (0,0) -- (0,0.25) -- (0.25,0.25)-- (0.25,0) -- (0,0); \end{tikzpicture} && q1\_rr &	std\_resp &	skew\_rr &	q3\_resp\\
        \begin{tikzpicture} \draw[draw=white, fill=green, opacity=0.2] (0,0) -- (0,0.25) -- (0.25,0.25)-- (0.25,0) -- (0,0); \end{tikzpicture} && med\_rr &	med\_rr &	respNK1 &	q1\_resp\\
         \Xhline{1pt}
         \begin{tikzpicture} \draw[draw=white, fill=green, opacity=1.0] (0,0) -- (0,0.25) -- (0.25,0.25)-- (0.25,0) -- (0,0); \end{tikzpicture}  & \multirow{5}{*}{\texttt{SpO}$_2$} & plethFFT& 	kurt\_pleth& 	med\_rr 	&plethTINT\\
        \begin{tikzpicture} \draw[draw=white, fill=green, opacity=0.8] (0,0) -- (0,0.25) -- (0.25,0.25)-- (0.25,0) -- (0,0); \end{tikzpicture} && plethINT &	plethFFT &	q1\_rr &	plethTNK1\\
        \begin{tikzpicture} \draw[draw=white, fill=green, opacity=0.6] (0,0) -- (0,0.25) -- (0.25,0.25)-- (0.25,0) -- (0,0); \end{tikzpicture} && plethNK1& 	plethINT &	q3\_rr& 	plethFFT\\
        \begin{tikzpicture} \draw[draw=white, fill=green, opacity=0.4] (0,0) -- (0,0.25) -- (0.25,0.25)-- (0.25,0) -- (0,0); \end{tikzpicture} && kurt\_pleth &	q3\_pleth &	plethNK1 &	plethTFFT\\
        \begin{tikzpicture} \draw[draw=white, fill=green, opacity=0.2] (0,0) -- (0,0.25) -- (0.25,0.25)-- (0.25,0) -- (0,0); \end{tikzpicture} && plethHeight &	plethNK1 &	skew\_pleth &	med\_SpO$_2$T\\
        \Xhline{1pt}
    \end{tabular}}
        \caption{Feature importance calculated for RR and SpO$_2$/SpO$_2$T alerts using \textit{Gini importance} (GI) and \textit{permutation feature importance} (PFI) are shown in decreasing order of importance. The ranked features between the weakly and fully supervised pipelines for both alert types show similarities and differences in the types of features used by the RF models for each pipeline.}
        \label{fig:featureimp}
\end{wrapfigure}

\paragraph{Performance Metrics}
For the RR alerts, the various performance metrics shown in Table \ref{tab:perfmetr2} highlight our WS model's surprising, but superior performance over its FS counterpart. On the other hand, analysis on the models' performance on SpO$_2$/SpO$_2$T alerts yielded more expected results, with the FS model performing slightly better than the WS. However, the WS model's performance is still noteworthy considering that the FS model had the immense advantage of accessing ground truth labels for training. The results also indicate that our models performed better at classifying RR alerts than SpO$_2$/SpO$_2$T alerts, consistent with prior work by Chen et al.\cite{chen}. However, the performance gap we found between the two alert types was slimmer, likely due to the inclusion of waveform data in our work.

\begin{figure}[!thb]
    \centering
  \begin{tabular}{cc}
    \includegraphics[width=0.19\textwidth]{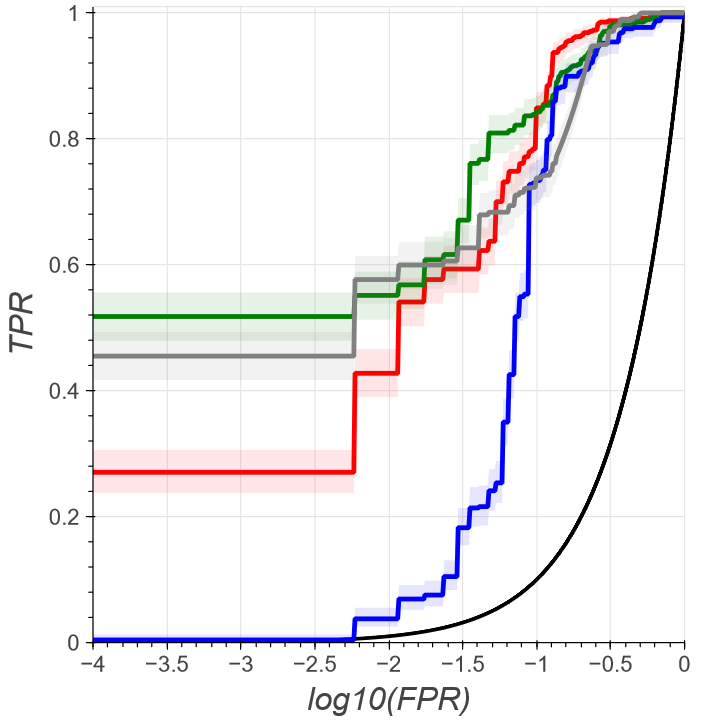} 
    &\includegraphics[width=0.19\textwidth]{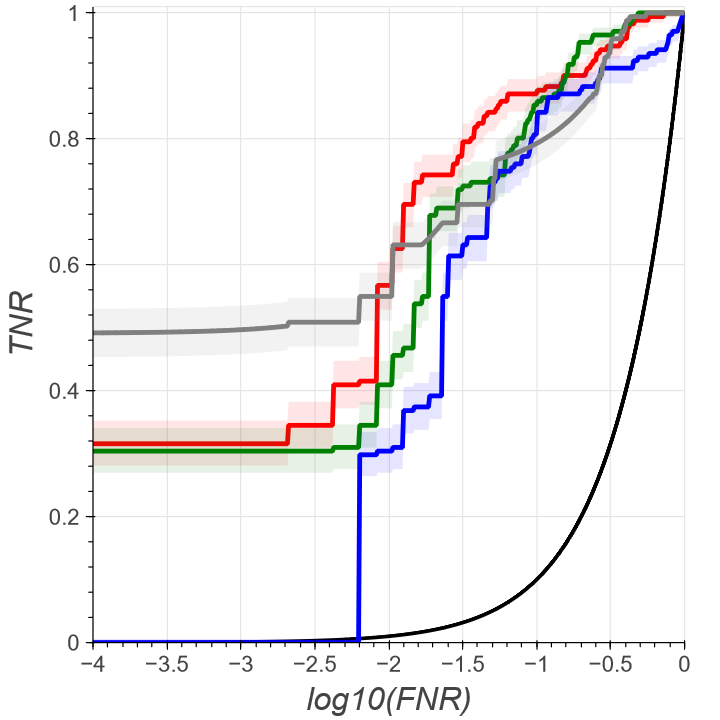}\\
    (i) & (ii) \\
  \end{tabular}
  \begin{tabular}{cc}
    \includegraphics[width=0.19\textwidth]{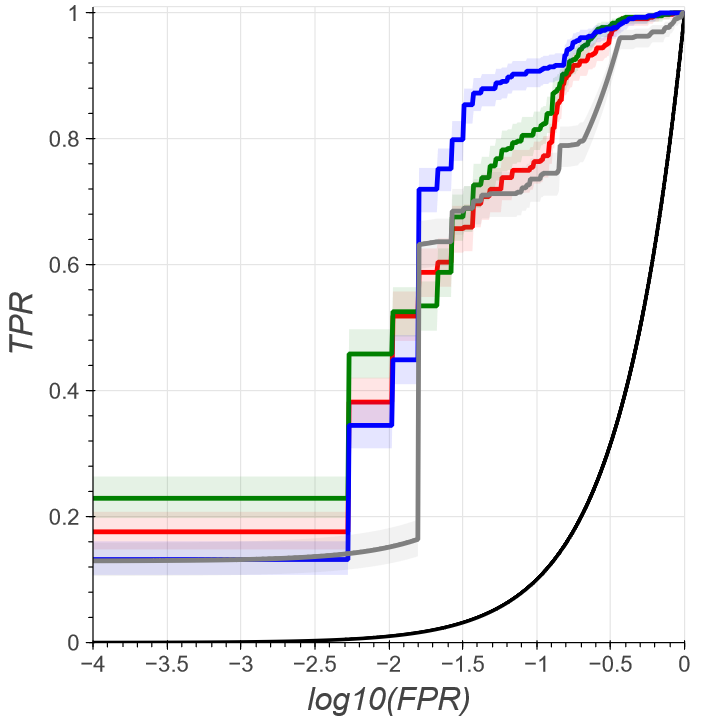}
    &\includegraphics[width=0.19\textwidth]{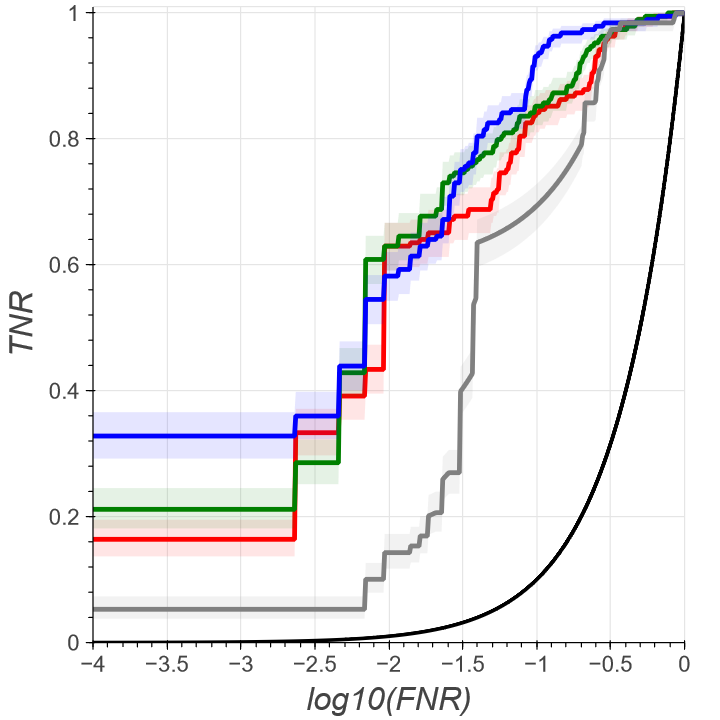}\\
    (iii) & (iv) \\
  \end{tabular}
    \caption{Log-scale ROC-AUC plots with 95\% Wilson score confidence intervals for RR (i \& ii) and SpO$_2$/SpO$_2$T alerts (iii \& iv), each with the WS (\textcolor{red}{\textit{red}}), majority labeler (\textcolor{green}{\textit{green}}), FS learning (\textcolor{blue}{\textit{blue}}), and probabilistic labels (\textcolor{grey}{\textit{grey}}). These plots highlight the WS pipeline's ability to keep up with the fully supervised method for the SpO$_2$/SpO$_2$T alerts, and outperform it on the RR alerts.}
    \label{fig:ROC}
\end{figure}

\begin{figure}[!thb]
    \noindent\centering
  \begin{tabular}{cc}
    \includegraphics[width=0.19\textwidth]{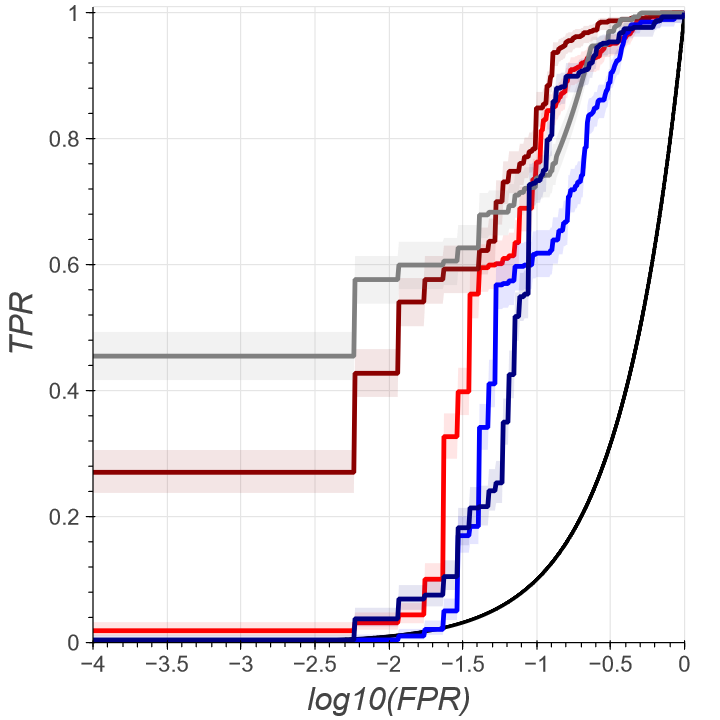} 
    &\includegraphics[width=0.19\textwidth]{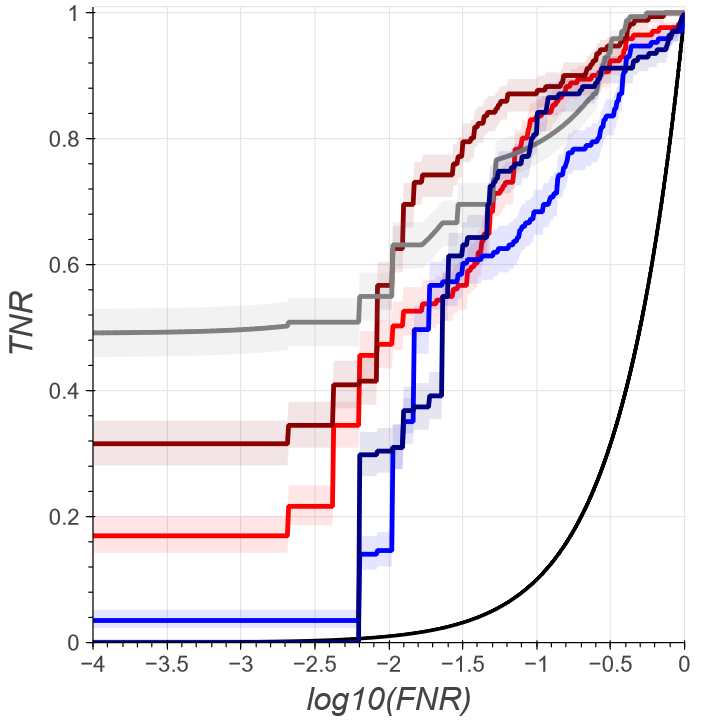}\\
    (i) & (ii) \\
  \end{tabular}
  \begin{tabular}{cc}
    \includegraphics[width=0.19\textwidth]{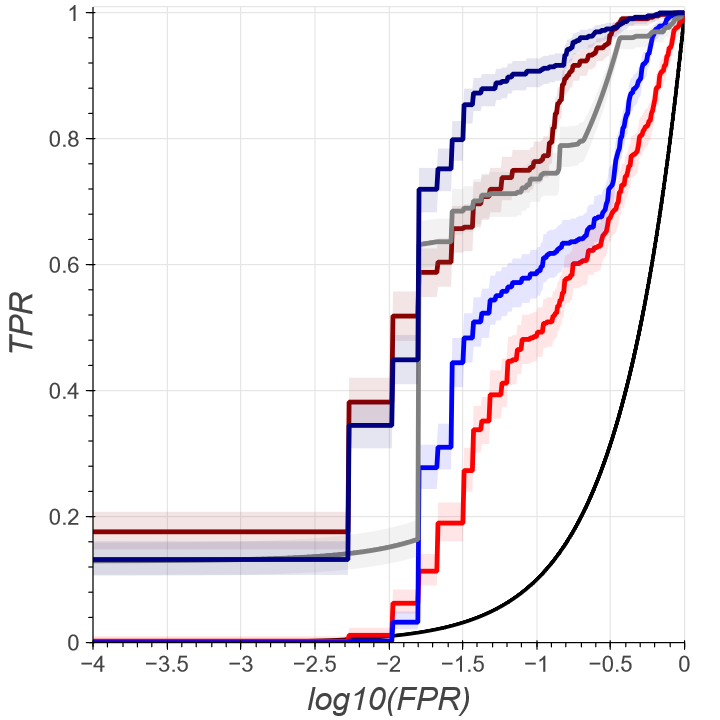}
    &\includegraphics[width=0.19\textwidth]{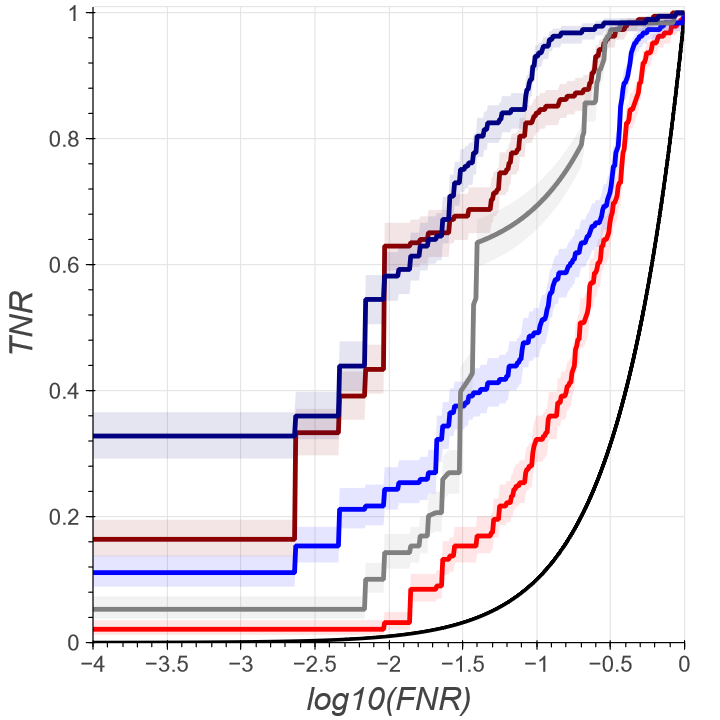}\\
    (iii) & (iv) \\
  \end{tabular}
    \caption{Log-scaled ROC plots with 95\% Wilson confidence intervals pertain for ablation experiments conducted on RR alerts (i \& ii) and SpO$_2$/SpO$_2$T alerts (iii \& iv). Each plot shows the WS pipeline \textit{without} waveform data (\textcolor{red}{\textit{red}}), \textit{with} waveform data (\textcolor{darkred}{\textit{dark-red}}), the FS pipeline without waveform data (\textcolor{blue}{\textit{blue}}), \textit{with} waveform data (\textcolor{darkblue}{\textit{dark-blue}}), and the probabilistic labels (\textcolor{grey}{\textit{grey}}). The separation between the curves indicate that waveform data is much more beneficial for oxygen saturation alerts than for RR alerts.
    }
    \label{fig:ROCABL}
\end{figure}

\paragraph{ROC Analysis}
In Figures \ref{fig:ROC} and \ref{fig:ROCABL} we plot pairs of Receiver Operating Characteristic (ROC) diagrams for each experimental configuration in logarithmic scale of the horizontal axis, to help focus the interpretation of the results on the low error rate settings which are of practical relevance in clinical decision support scenarios. One plot in each pair shows true positive rate as a function of logarithmically scaled false positive rate, the other shows the other end of the ROC plot by presenting the same data in the coordinates of true negative rate as a function of logarithmically scaled false negative rate. Each plot includes a random model line in solid dark grey for viewers' reference

From ROC plots for RR alerts shown in Figure ~\ref{fig:ROC} (i \& ii), it is clear that our WS model has a higher TPR and TNR at nearly every FPR and FNR setting, respectively. For SpO$_2$/SpO$_2$T (Plots iii \& iv in Figure~\ref{fig:ROC}), that is not the case. Nonetheless, WS still performs comparably to FS, despite not having access to ground truth labels, underscoring the impressive capabilities of weak supervision as applied to VS alert classification.

\paragraph{Answers to Additional Research Questions}
\textit{Our weakly and fully supervised RF models are learning similar patterns.} We found considerable overlap between important features for our weakly and fully supervised RF classifiers, despite some minor differences in the feature importance ranking. 
For example for RR alerts, our models found 
the standard deviation (\texttt{std\_resp}) and the height of the \texttt{resp} waveform (\texttt{respHeight}) to be the most important. For SpO$_2$/SpO$_2$T alerts, we found the \texttt{HR} derived from the primary harmonic of the \texttt{pleth} (\texttt{plethFFT}), and by counting its peaks (\textit{plethINT})\footnote{Specifically, \textit{plethINT} is based on the number of common peaks found using the peak finding functions of \texttt{sciPy} and \texttt{Neurokit2}.} to be high-ranking across both models in terms of GI and PFI. The apparent discrepancies in rankings may be due to the different ways in which GI and PFI compute feature importance. Nevertheless, the large overlap in high-ranking features for both alert types, across both types of models, and for both feature importance metrics, indicates that the weakly and fully supervised RF models may be learning similar patterns for both VS alert types.

\textit{Waveform data is helpful for RR alerts, but almost essential for SpO$_2$/SpO$_2$T alerts.} The log-scale ROC plots in Figure \ref{fig:ROCABL} neatly visualize the predictive utility of waveform data for both the RR and SpO$_2$/SpO$_2$T alerts. For RR alerts, the plots show some separation between models with and without access to waveform data, indicating the slight usefulness of waveform data for RR alert classification. In contrast, the plots for oxygen saturation alerts show a much larger gap, with models having access to waveform data performing much better than those without. The significant predictive utility of waveform data for oxygen saturation alert classification is further substantiated by the ubiquity of waveform features in the top echelon of feature importance rankings, as highlighted previously and shown in Table~\ref{fig:featureimp}. 


\section*{Discussion}
This work has five main takeaways: (1)~The novel core methodologies we developed to derive VS numeric values from time series waveform data were reliable, and have meaningful applications beyond the scope of this project. (2)~Both the fully and weakly supervised pipelines - when validated on unseen data from a unique patient - remained robust and performed well, with AUC values ranging from 0.898 to 0.964 for all the models. (3)~The predictive utility of waveform data was found to be only slight for RR alerts, but significantly important for SpO$_2$/SpO$_2$T alerts. (4)~The WS models were shown to perform on par with their FS counterparts, and for the RR alerts, even outperform them. (5)~Perhaps most importantly, the WS models could be built within a span of a few hours, and without the significant involvement or time of domain experts, streamlining the process of building accurate and efficient VS alert classifiers. Overall, this work demonstrates the efficacy of weak supervision as a framework to streamline the process for building ML models for HC applications in a scalable fashion, contributing to a broadening of the social impact of powerful machine learning methodologies.

\section*{Limitations \& Future Work}
There are a few limitations of this work. Firstly, it assumes a priori knowledge of approximate real versus artifact class balances of vital sign alerts. However, domain experts often already have this knowledge, so these models can still be built to accomplish their goals. 
Secondly, due to the design of our study, the WS model is currently best used as a ``fact-checker” that lends a secondary opinion on archived vital sign alert data. In the future, analysis should be carried out to measure the speed and latency of the classification algorithm, before eventually optimizing the design to create and implement a real-time artifact alert adjudication system. Despite these limitations, the promising results indicate that a trained WS model could eventually serve as an effective tool for medical practitioners to combat alarm fatigue in the acute and intensive care settings.

\section*{Acknowledgements}
This work was partially supported by a fellowship from Carnegie Mellon University’s Center for Machine Learning and Health to M.G.

\makeatletter
\makeatother

\bibliographystyle{unsrt}
\bibliography{bibliography}

\end{document}